# Iranis: A Large-scale Dataset of Farsi License Plate Characters

Ali Tourani
*Department of Computer Engineering, University of Guilan,* Rasht, Iran
tourani@msc.guilan.ac.ir

Sajjad Soroori
*Department of Computer Engineering, University of Guilan,* Rasht, Iran
sajjad_soroori@msc.guilan.ac.ir

Asadollah Shahbahrami
*Department of Computer Engineering, University of Guilan,* Rasht, Iran
shahbahrami@guilan.ac.ir

Alireza Akoushideh
*Shahid-Chamran College Technical and Vocational University, Guilan branch,* Rasht, Iran
akushide@tvu.ac.ir

*Abstract*— Providing huge amounts of data is a fundamental demand when dealing with Deep Neural Networks (DNNs). Employing these algorithms to solve computer vision problems resulted in the advent of various image datasets to feed the most common visual imagery deep structures, known as Convolutional Neural Networks (CNNs). In this regard, some datasets can be found that contain hundreds or even thousands of images for license plate detection and optical character recognition purposes. However, no publicly available image dataset provides such data for the recognition of Farsi characters used in car license plates. The gap has to be filled due to the numerous advantages of developing accurate deep learning-based systems for law enforcement and surveillance purposes. This paper introduces a large-scale dataset that includes images of numbers and characters used in Iranian car license plates. The dataset, named *Iranis*, contains more than 83,000 images of Farsi numbers and letters collected from real-world license plate images captured by various cameras. The variety of instances in terms of camera shooting angle, illumination, resolution, and contrast make the dataset a proper choice for training DNNs. Dataset images are manually annotated for object detection and image classification. Finally, and to build a baseline for Farsi character recognition, the paper provides a performance analysis using a YOLO v.3 object detector.

## I. Introduction

Intelligent Transportation Systems (ITS) require precise and novel tools for urban traffic management and control. In this regard, cameras are considered as one of the best solutions to provide real-time data for surveillance and security, traffic congestion management, and law enforcement [1]. Employing computer vision techniques has provided the advantage of automatic information retrieval from still images and videos captured by the cameras [2]. Since deep learning approaches have shown superb outcomes and robust performance when dealing with classification challenges, they have been applied to computer vision applications in recent years [3]. Convolutional Neural Networks (CNNs), Deep Belief Networks (DBN), and Autoencoders are some of the well-known deep learning schemes used in the mentioned field [3-6]. Various applications of ITS, such as traffic estimation, automatic license plate detection, and vehicle speed measurement have been developed using the mentioned approaches in recent years [2]. In this respect, localization of vehicle license plates and extracting the characters inside them can assist authorities in their enforcement of traffic laws.

*This work was supported by the Technology Incubation Center of the University of Guilan under grant UOG-IC-303,15/20493, date 2019-02-18, and DadeKavan Khazar Pouya company.*

There are two main layouts for vehicle license plates in Iran: *typical* and *special*. In the typical layout, the license plate includes two partitions, where the left part holds a string of numbers and letters, and the right part refers to a two-digit number that indicates the issuer province code. In this regard, the combination of the two parts shapes a unique identifier for each vehicle. On the other hand, the special layout devotes to the free trade zone vehicles. In contrast with the typical form, in which the license plates have European standard dimensions, the special type has a completely different width-to-height ratio. Furthermore, the plate is black on white and a blue section in the leftmost part holds the free trade zone's logo, *i.e.* Anzali, Kish, Arvand, etc. It should be noted that in the newest design, the logo has been replaced by a two-digit code for each trade zone. The license plates in this category contain no letters and only five numbers along with their equivalent English are provided. Fig. 1 demonstrates the mentioned layouts in brief.

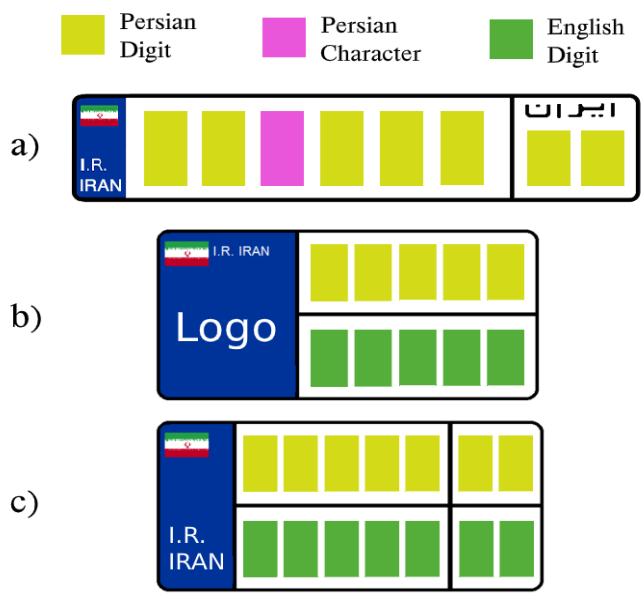

Fig. 1. Layouts of Iranian car license plates: a) typical, b) special, and c) new design of the special layout.

The Farsi -*i.e. Farsi/Arabic*- letter in the typical license plate is used to distinguish the type of the plate. It should be noted that for the private cars of people with disabilities, there is a wheelchair symbol instead of the character. Additionally, the background of these plates may vary regarding the type and usage of the vehicle. For instance, white is used for private cars, while yellow indicates taxis and public vehicles. These variations make the license plate detection and character recognition process a challenging task. Fig. 2 shows different types of car license plates in Iran.

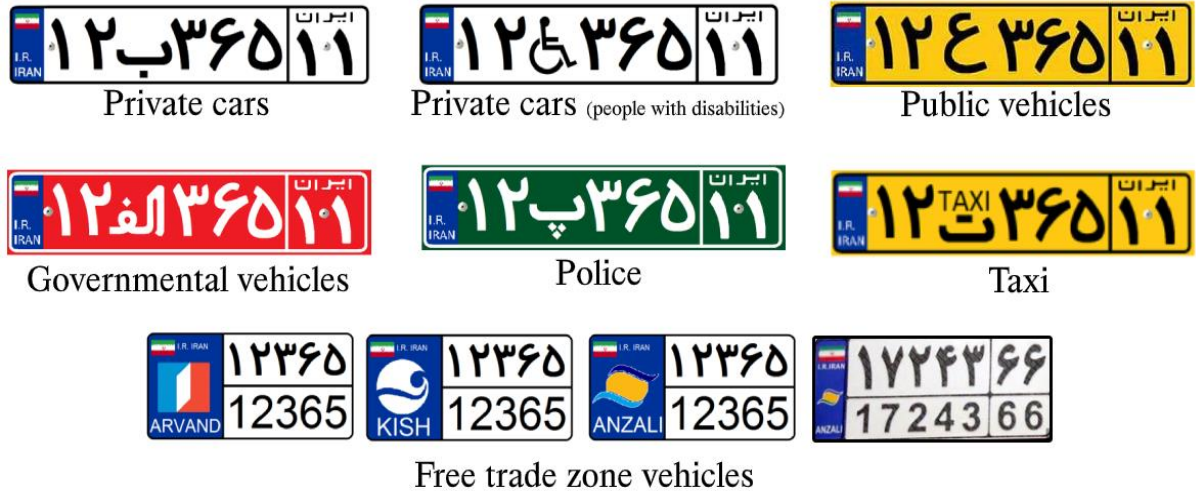
Fig. 2. Variations of Iranianc car license plates [8].

According to Fig. 1 and Fig. 2, the variety of layouts in Iranian car license plates and the similarity of some letters and numbers in the Farsi alphabet cause challenges for Automatic License Plate Detection and Recognition (ALPR) purposes. ALPR plays a key role in the functionality of camera-based ITS and smart cities, as various accurate and reliable systems are highly dependent on the correct recognition of characters of the license plates. In this regard, providing real-world data can be extremely vital for developing state-of-the-art applications, especially machine learning and deep learning approaches.

In this paper, a large-scale dataset named *Iranis*, containing the images of Farsi characters is introduced. The authors believe that the mentioned dataset can deal with the lack of proper Farsi letters and numbers dataset for training robust ALPR applications. In this regard, the main contributions of the paper include building a multimodal, stable dataset for object detection and character recognition in Iranian car license plates and indicating its application in ALPR.

The rest of the paper is organized as follows: Section II describes the motivation of the paper and presents some related works. The introduction and discussion of the proposed dataset are accessible in Section III. In Section IV, we evaluate the functionality of the proposed dataset using a deep learning scheme. The paper ends with some conclusions in Section V.

## II. MOTIVATION AND RELATED WORKS

Since collecting, labeling, and classifying visual data is a tough process and requires much effort and time, there has always been a pale interest among researchers to develop large-scale datasets for computer vision and machine learning. Although there are some standard datasets available for license plate detection [9-12] and character recognition [13-15], they cannot be employed to recognize Farsi characters in images.

The authors of this paper have contributed to many computer vision and machine learning projects and proposed various approaches in the field of ITS. Some of these approaches include vehicle count using video processing [16], deep learning-based vehicle detection [17], vehicle speed measurement [18-19], license plate localization [8, 20], and Farsi character recognition [8]. Accordingly, we claim that we have felt the essence of reliable data for the development of domestic robust applications for

ITS. This motivated us to make the proposed dataset and make it available for enthusiastic researchers.

To the best of our analysis, there is no publicly available Farsi characters dataset extracted from real-world Iranian license plate images with available annotation data for classification. Although some works were performed by various researchers [21-24], they have been developed for handwritten Optical Character Recognition (OCR) researches and text-processing applications. These datasets cannot be used to recognize license plate characters due to variations of handwriting styles. In contrast with the mentioned approaches, a dataset of Farsi license plate characters is introduced in [25] in which the letters and numbers are directly extracted from real-world license plate images. Although it covers all possible characters, the number of instances for each letter fluctuates between 200 and 580. Another similar dataset is introduced in [26] which suffers a lack of data and no object detection annotation is available for the instances. The mentioned features make it inappropriate for classification and deep learning approaches, where the variety of data samples plays a vital role.

To cover the mentioned drawbacks, Iranis is developed to address the necessity of data requirement for Iranian ALPR systems and fill the mentioned gap to a large extent. It covers all commonplace variations of Farsi license plate characters and provides annotation data for classification.

### III. DATASET DESCRIPTION

Iranis is a large-scale image set, collected from real-world license plate images. These images were captured by uncalibrated cameras under various illumination conditions and may vary in resolution, contrast, shooting angle, and distance. In this regard, the processes of image collection, filtering, labeling, and bounding-box drawing -*i.e. image annotation*- were accomplished by a group of researchers at the University of Guilan, Iran. Consequently, the dataset contains a wide range of samples for each letter or number used in Iranian car license plates. It should be noted that since license plates are used for the identification of individuals, we do not make the images of the license plates publicly available regarding the data protection and privacy law.

To obtain characters from license plates, we have proposed a two-stage deep learning approach that localizes plates in images, and then segments and extracts the inside texts. Further information about these stages can be found in our recently published work [8]. Table I illustrates the characteristics of the dataset in detail. It should be noted that the data annotation process was accomplished through an automatic process with supervised validation. To label data, we utilized Ybat [28], a YOLO bounding box annotation tool. In this step, a group of graduate and undergraduate students were asked to label each letter or number found in the license plate area and ignore the characters that are unclear and unable to be recognized. To ensure the characters were correctly labeled and avoid any potential misclassifications, labels were double-checked by different students. Finally, the outputs were saved as JSON format for classification and validation.

TABLE I. INSTANCES OF IRANIS IMAGE SET

| Type | Label (class name) | Corresponding character, digit, or symbol | Number of instances |
|---|---|---|---|
| Number | 0 | ٠ | 2501 |
| | 1 | ١ | 3495 |
| | 2 | ٢ | 3930 |
| | 3 | ٣ | 2745 |
| | 4 | ۴ | 5774 |
| | 5 | ۵ | 3610 |
| | 6 | ۶ | 5753 |
| | 7 | ٧ | 3736 |
| | 8 | ٨ | 3583 |
| | 9 | ٩ | 3528 |
| Letter | A | الف | 2517 |
| | B | ب | 2511 |
| | P | پ | 2519 |
| | J | ج | 2505 |
| | H | ح | 2558 |
| | D | د | 2504 |
| | Sin | س | 2445 |
| | Sad | ص | 2515 |
| | T | ط | 2512 |
| | Gh | ق | 2482 |
| | L | ل | 2502 |
| | M | م | 2500 |
| | N | ن | 2558 |
| | V | و | 2509 |
| | Y | ى | 2491 |
| | PuV | ع | 2508 |
| | Taxi | ت | 2551 |
| Symbol | PwD | ﻉ | 2502 |
| **Total** | | | **83,844** |

According to Table I, there are more than 38,000 instances of Farsi "numbers" with the numerical labels 0 to 9 in the dataset, while the number of Farsi letters is greater than 42,000. Hereby, Iranis contains 28 classes, including ten numbers, seventeen letters, and one symbol. Fig. 3 shows some instances of the dataset along with their corresponding class names. Additionally, some notes should be considered about the dataset instances:

- The dataset does not contain the English numbers used in free trade zone plates,
- Motorcycle license plates have a completely different layout, font size, and dimensions, and have not been included in the proposed dataset,
- We have ignored some extremely rare letters in data classification, like characters "ک" for agricultural cars, "ز" for the ministry of defense vehicles, and "D" for diplomatic and consular corps vehicles,
- Some augmentation techniques such as multiply, rotation, and additive Gaussian noise were used according to [27] to increase the number of remaining scarce samples, like classes *P* for police and *PwD* for private cars of people with disabilities. The outcomes of the augmentation process are available in [8].
- Likewise, the number zero with class name *0*, is only appeared in the rightmost place of the issuer province code section -*e.g. 10, 20, and 70*- and hence, the instances were scarce and have been augmented.

| | | | | | | | |
|---|---|---|---|---|---|---|---|
| 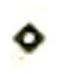 | 0 (zero) | 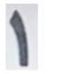 | 1 (one) | 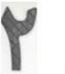 | 2 (two) | 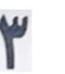 | 3 (three) |
| 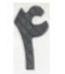 | 4 (four) | 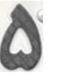 | 5 (five) | 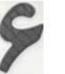 | 6 (six) | 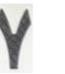 | 7 (seven) |
| 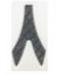 | 8 (eight) | 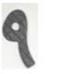 | 9 (nine) | 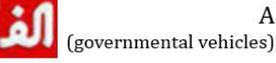 | A (governmental vehicles) | 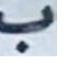 | B |
| 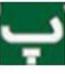 | P (police) | 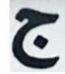 | J | 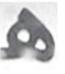 | H | 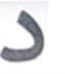 | D |
| 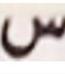 | Sin | 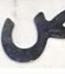 | Sad | 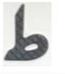 | T | 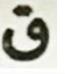 | Gh |
| 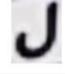 | L | 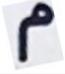 | M | 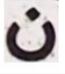 | N | 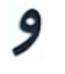 | V |
| 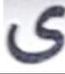 | Y | 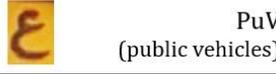 | PuV (public vehicles) | 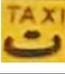 | Taxi | 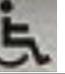 | PwD (people with disabilities) |

Fig. 3. Sample instances available in Iranis image set.

However, as Fig. 3 shows, Iranis covers almost all commonplace letters found in license plates along with all variations of numbers used in both typical and special types. The ignored classes mentioned above can rarely be seen in urban areas, which makes the dataset appropriate for practical usage.

IV. BENCHMARK

To assess the applicability of our dataset, we have used it to train a YOLO version 3 [29] network. The network was implemented using Python 3, TensorFlow framework, and CUDNN. The train and test processes were accomplished on a machine with an Intel Core i7 processor, equipped with 48 Gigabytes of RAM and an 8 Gigabytes Nvidia GPU. We have presented the training process and experimental results are in our recent publication in [8].

To employ the proposed dataset, we have divided the data into train-set and test-set, containing 85% and 15% of data, respectively. Table II illustrates the experimental results of applying the proposed

YOLO network on Iranis dataset. The training process of the network took 45 epochs regarding the utilized hardware, and loss parameter did not increase from 2.46 after that.

TABLE II. UTILIZING YOLO VERSION 3 AS A BENCHMARK FOR TEST.

| Classes | # of data | | Precision | Recall |
| --- | --- | --- | --- | --- |
| | Train set | Test set | | |
| *Number* | 32,857 | 5,798 | 0.965717 | 0.980368 |
| *Letter* | 36,284 | 6,403 | 0.982975 | 0.981258 |
| *Symbol* | 2,127 | 375 | 0.975676 | 0.986339 |
| **Total** | **71,268** | **12,576** | **0.974804** | **0.981003** |

According to Table II, the precision and recall of the network proved that our dataset can be employed for practical applications of ALPR. It provides a wide range of data instances along with their corresponding bounding boxes for the training process, which makes Iranis a proper choice for Farsi character recognition applications. Fig. 4 shows the impact of employing the object detector on our dataset.

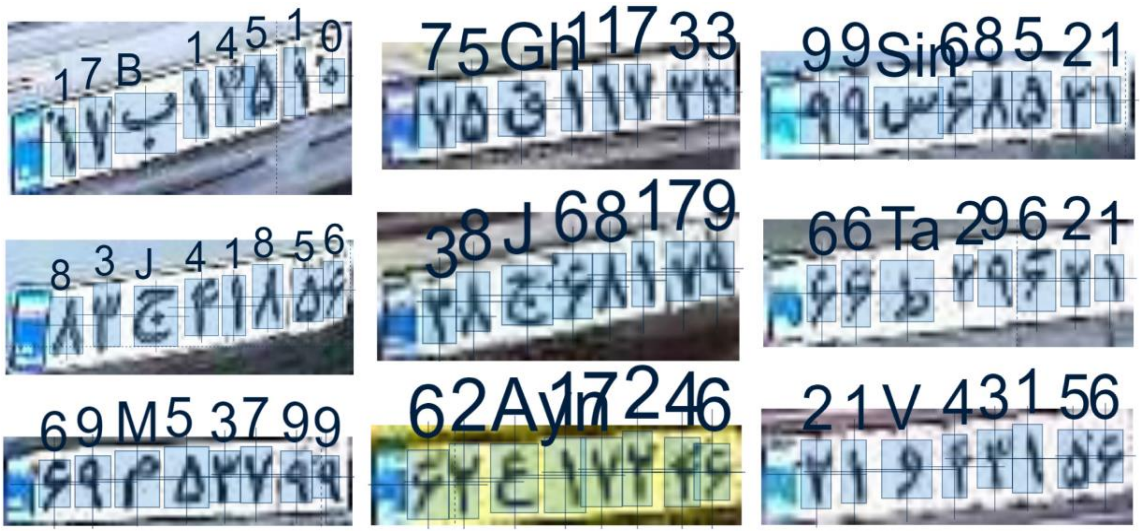

Fig. 4. Sample outcomes of applying YOLO v.3 on the dataset [8].

The dataset is publicly available for academic usage on top sources for machine learning datasets. As for future works, our major goal is to focus on motorcycle license plates and collect more data to cover even the scarce letters that have been ignored for now. The updates will be available on the mentioned web page once in a while. In this regard, the authors encourage any contributions to improve the quality of the dataset.

## V. Conclusions

Intelligent Transportation Systems play a key role in the development of smart cities, regarding their impacts on traffic and congestion control. One of the most important systems used in ITS is the automatic detection and recognition of vehicle license plates. To develop accurate and robust systems in this field, providing real-world data of the license plate images and letters used in the plates is vital. Considering the importance of data and since there is no available dataset with the mentioned characteristics that contain Farsi characters used in Iranian license plates, this paper introduced a large-scale dataset for Automatic License Plate Recognition applications. Iranis is a publicly available dataset and can be used by all computer vision researchers. The dataset contains more than 83,000 instances of letters and numbers that are classified into 28 classes, makes it a proper choice for Deep Learning purposes. In this regard, the annotation data, i.e. bounding boxes, of all instances is available for train and test stages. Since the instances of the dataset are cropped from real-world Iranian car license plates, training a model using this dataset is highly recommended for practical usage. The dataset may enlarge and improve overtime to cover potential issues and defects.

## Acknowledgment

The authors would like to thank the Technology Incubation Center of the University of Guilan, Guilan Science and Technology Park, and DadeKavan Khazar Pouya company for their kind supports. We also thank all the graduate and undergraduate students who contributed and were involved in acquiring and labeling data.